\documentclass[runningheads]{llncs}

\usepackage{graphicx}
\usepackage{amssymb,amssymb} 
\usepackage{color}
\usepackage{hyperref}

\usepackage{amsmath}
\usepackage{amssymb}
\usepackage{bm}
\usepackage{xspace}
\usepackage{booktabs}
\usepackage{multirow}
\usepackage{makecell}
\usepackage{pifont}
\usepackage{subfig}
\usepackage{bbold}

\newcommand{\GW}{GuessWhat?!\xspace}
\newcommand{\Referit}{ReferIt\xspace}
\newcommand{\memfilm}{Multi-hop FiLM Generator\xspace}
\newcommand{\sfilm}{Single-hop FiLM\xspace}
\newcommand{\memfilmshort}{Multi-hop FiLM\xspace}
\newcommand{\Image}{\mathcal{I}}

\newcolumntype{x}[1]{>{\centering\arraybackslash}p{#1}}
\newcommand{\argmax}{\mathop{\mathrm{argmax}}}          

\newcommand{\cmark}{\textcolor{green}{\ding{51}}}
\newcommand{\xmark}{\textcolor{red}{\ding{55}}}

\begin{document}
\pagestyle{headings}
\mainmatter
\def\ECCV18SubNumber{2093}  

\title{Visual Reasoning with Multi-hop Feature Modulation} 

\titlerunning{Visual Reasoning with Multi-hop Feature Modulation}

\authorrunning{F. Strub and M. Seurin and  E. Perez and  H. Vries et al. }

\author{
Florian Strub\textnormal{\textsuperscript{1}},
Mathieu Seurin\textnormal{\textsuperscript{1}}, 
Ethan Perez\textnormal{\textsuperscript{2,3}},
Harm de Vries\textnormal{\textsuperscript{2}},\\
J\'er\'emie Mary \textnormal{\textsuperscript{4}},
Philippe Preux \textnormal{\textsuperscript{1}},
Aaron Courville\textnormal{\textsuperscript{2,5}}
Olivier Pietquin\textnormal{\textsuperscript{6}}\\
}

\institute{
\textsuperscript{1}Univ. Lille, CNRS, Inria, UMR 9189 CRIStAL 
\textsuperscript{2}MILA, Universit\'e de Montr\'eal,\\
\textsuperscript{3}Rice University,
\textsuperscript{4}Criteo,
\textsuperscript{5}CIFAR Fellow,
\textsuperscript{6}Google Brain
}

\maketitle
\begin{abstract}
Recent breakthroughs in computer vision and natural language processing have spurred interest in challenging multi-modal tasks such as visual question-answering and visual dialogue. For such tasks, one successful approach is to condition image-based convolutional network computation on language via Feature-wise Linear Modulation (FiLM) layers, i.e.\@, per-channel scaling and shifting.
We propose to generate the parameters of FiLM layers going up the hierarchy of a convolutional network in a multi-hop fashion rather than all at once, as in prior work.
By alternating between attending to the language input and generating FiLM layer parameters, this approach is better able to scale to settings with longer input sequences such as dialogue.
We demonstrate that multi-hop FiLM generation achieves state-of-the-art for the short input sequence task \Referit --- on-par with single-hop FiLM generation --- while also significantly outperforming prior state-of-the-art and single-hop FiLM generation on the \GW visual dialogue task.


\end{abstract}

\section{Introduction}


Computer vision has witnessed many impressive breakthroughs over the past decades in image classification~\cite{krizhevsky2012imagenet,he2016deep}, image segmentation~\cite{long2015fully}, and object detection~\cite{girshick2014rich} by applying convolutional neural networks to large-scale, labeled datasets, often exceeding human performance. These systems give outputs such as class labels, segmentation masks, or bounding boxes, but it would be more natural for humans to interact with these systems through natural language. To this end, the research community has introduced various multi-modal tasks, such as image captioning~\cite{xu2015show}, referring expressions~\cite{kazemzadeh2014referitgame}, visual question-answering~\cite{antol2015vqa,malinowski2015ask}, visual reasoning~\cite{johnson2017clevr}, and visual dialogue~\cite{de2017guesswhat,das2017visual}.

These tasks require models to effectively integrate information from both vision and language.
One common approach is to process both modalities independently with large unimodal networks before combining them through concatenation~\cite{malinowski2015ask}, element-wise product~\cite{kim2016multimodal,lu2016hierarchical}, or bilinear pooling~\cite{fukui2016multimodal}.  Inspired by the success of attention in machine translation~\cite{bahdanau2014neural}, several works have proposed to incorporate various forms of spatial attention to bias models towards focusing on question-specific image regions~\cite{xu2015show,xu2016ask}. However, spatial attention sometimes only gives modest improvements over simple baselines for visual question answering~\cite{jabri16vqa} and can struggle on questions involving multi-step reasoning~\cite{johnson2017clevr}.

More recently,~\cite{devries2017modulating,perez2018film} introduced Feature-wise Linear Modulation (FiLM) layers as a promising approach for vision-and-language tasks. These layers apply a per-channel scaling and shifting to a convolutional network's visual features, conditioned on an external input such as language, e.g.\@, captions, questions, or full dialogues. Such feature-wise affine transformations allow models to dynamically highlight the key visual features for the task at hand. The parameters of FiLM layers which scale and shift features or feature maps are determined by a separate network, the so-called \emph{FiLM generator}, which predicts these parameters using the external conditioning input. Within various architectures, FiLM has outperformed prior state-of-art for visual question-answering~\cite{devries2017modulating,perez2018film}, multi-modal translation~\cite{delbrouck2017modulating}, and language-guided image segmentation~\cite{rupprecht2018guide}.

\begin{figure}[t]
\begin{minipage}{0.30\textwidth}
\centering
\includegraphics[width=0.85\textwidth]{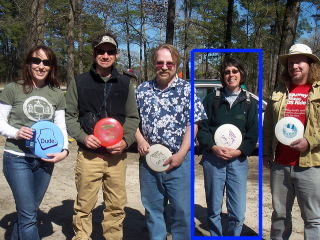}
\end{minipage}
\begin{minipage}{0.68\textwidth}
\begin{tabular}{llll}
\toprule
\textit{\Referit} & \quad\quad & \textit{\GW} & \\
\midrule
- The girl with a sweater & & Is it a person? &\quad Yes\\
- The fourth person & & Is it a girl? &\quad Yes\\
- The girl holding a white & & Does she have a blue&\quad No\\
frisbee & & frisbee? &\\
\end{tabular}
\end{minipage}
\caption{The \Referit task identifies a selected object (in the bounding box) using a single expression, while in \GW, a speaker localizes the object with a series of yes or no questions.}
\label{fig:example}
\end{figure}

However, the best way to design the FiLM generator is still an open question. For visual question-answering and visual reasoning, prior work uses single-hop FiLM generators that predict all FiLM parameters at once~\cite{perez2018film,devries2017modulating}. That is, a Recurrent Neural Network (RNN) sequentially processes input language tokens and then outputs all FiLM parameters via a Multi-Layer Perceptron (MLP). In this paper, we argue that using a \textit{Multi-hop FiLM Generator} is better suited for tasks involving longer input sequences and multi-step reasoning such as dialogue. Even for shorter input sequence tasks, single-hop FiLM generators can require a large RNN to achieve strong performance; on the CLEVR visual reasoning task~\cite{johnson2017clevr} which only involves a small vocabulary and templated questions, the FiLM generator in~\cite{perez2018film} uses an RNN with 4096 hidden units that comprises almost 90\% of the model's parameters. Models with Multi-hop FiLM Generators may thus be easier to scale to more difficult tasks involving human-generated language involving larger vocabularies and more ambiguity.


As an intuitive example, consider the dialogue in Fig.~\ref{fig:example} through which one speaker localizes the second girl in the image, the one who does not ``have a blue frisbee.'' For this task, a single-hop model must determine upfront what steps of reasoning to carry out over the image and in what order; thus, it might decide in a single shot to highlight feature maps throughout the visual network detecting either non-blue colors or girls.
In contrast, a multi-hop model may first determine the most immediate step of reasoning necessary (i.e.\@, locate the girls), highlight the relevant visual features, and then determine the next immediate step of reasoning necessary (i.e.\@, locate the blue frisbee), and so on.
While it may be appropriate to reason in either way, the latter approach may scale better to longer language inputs and/or or to ambiguous images where the full sequence of reasoning steps is hard to determine upfront, which can even be further enhanced by having intermediate feedback while processing the image. 

In this paper, we therefore explore several approaches to generating FiLM parameters in multiple hops. These approaches introduce an intermediate context embedding that controls the language and visual processing, and they alternate between updating the context embedding via an attention mechanism over the language sequence (and optionally by incorporating image activations) and predicting the FiLM parameters. We evaluate \memfilmshort generation on \Referit~\cite{kazemzadeh2014referitgame} and \GW~\cite{de2017guesswhat}, two vision-and-language tasks illustrated in Fig.~\ref{fig:example}. We show that \memfilmshort models significantly outperform their single-hop counterparts and prior state-of-the-art for the longer input sequence, dialogue-based \GW task while matching the state-of-the-art performance of other models on \Referit. Our best \GW model only updates the context embedding using the language input, while for \Referit, incorporating visual feedback to update the context embedding improves performance.

\noindent In summary, this paper makes the following contributions: 
\begin{itemize}
\item We introduce the \memfilmshort architecture and demonstrate that our approach matches or significantly improves state-of-the-art on the \GW Oracle task, \GW Guesser task, and \Referit Guesser task.
\item We show \memfilmshort models outperforms their single-hop counterparts on vision-and-language tasks involving complex visual reasoning.
\item We find that updating the context embedding of \memfilm based on visual feedback may be helpful in some cases, such as  for tasks which do not include object category labels like \Referit.
\end{itemize}

\section{Background}
In this section, we explain the prerequisites to understanding our model: RNNs, attention mechanisms, and FiLM. We subsequently use these building blocks to propose a \memfilmshort model.

\subsection{Recurrent Neural Networks}
One common approach in natural language processing is to use a Recurrent Neural Network (RNN) to encode some linguistic input sequence $l$ into a fixed-size embedding. The input (such as a question or dialogue) consists of a sequence of words $\omega_{1:T}$ of length $T$, where each word $\omega_t$ is contained within a predefined vocabulary $\mathcal{V}$. We embed each input token via a learned look-up table $e$ and obtain a dense word-embedding $\bm{e}_{\omega_t} = e(\omega_t)$.
The sequence of embeddings $\{\bm{e}_{\omega_t}\}_{t=1}^T$ is then fed to a RNN, which produces a sequence of hidden states $\{\bm{s}_{t}\}_{t=1}^T$ by repeatedly applying a transition function $f$: $
\bm{s}_{t+1} = f(\bm{s}_{t}, \bm{e}_{\omega_t})
$
To better handle long-term dependencies in the input sequence, we use a Gated Recurrent Unit (GRU)~\cite{chung2014empirical} with layer normalization~\cite{ba2016layer} as transition function. In this work, we use a bidirectional GRU, which consists of one forward GRU, producing hidden states  $\overrightarrow{\bm{s}_{t}}$ by running from $\omega_1$ to $\omega_T$, and a second backward GRU, producing states $\overleftarrow{\bm{s}_{t}}$ by running from $\omega_T$ to $\omega_1$. We concatenate both unidirectional GRU states $\bm{s}_{t} = [ \overrightarrow{\bm{s}_{t}}; \overleftarrow{\bm{s}_{t}} ]$ at each step $t$ to get a final GRU state, which we then use as the compressed embedding $\bm{e}_l$ of the linguistic sequence $l$. 

\subsection{Attention}
The form of attention we consider was first introduced in the context of machine translation~\cite{bahdanau2014neural,luong2015effective}. This mechanism takes a weighted average of the hidden states of an encoding RNN based on their relevance to a decoding RNN at various decoding time steps. Subsequent \textit{spatial} attention mechanisms have extended the original mechanism to image captioning~\cite{xu2015show} and other vision-and-language tasks~\cite{xu2016ask,kim2016hadamard}.
More formally, given an arbitrary linguistic embedding $\bm{e}_l$ and image activations $\bm{F}_{w,h,c}$ where $w$, $h$, $c$ are the width, height, and channel indices, respectively, of the image features $\bm{F}$ at one layer, we obtain a final visual embedding $\bm{e}_v$ as follows:
\begin{align}
	\small
	\xi_{w, h} = MLP(g(\bm{F_{w,h,\cdot}},\bm{e_l})) \;;\quad
    \alpha_{w, h} = \frac{\exp(\xi_{w, h})}{\sum_{w',h'} \exp(\xi_{w',h'})} \quad;\;
    \bm{e}_{v} =  \sum_{w, h} \alpha_{w, h} \bm{F_{w,h,\cdot}},
\end{align}
where $MLP$ is a multi-layer perceptron and $g(.,.)$ is an arbitrary fusion mechanism (concatenation, element-wise product, etc.). We will use Multi-modal Low-rank Bilinear (MLB) attention~\cite{kim2016hadamard} which defines $g(.,.)$ as:
\begin{equation}
    g(\bm{F}_{w,h,\cdot}, \bm{e_l}) = \tanh(\bm{U}^{T}\bm{F}_{w, h, \cdot}) \circ \tanh(\bm{V}^{T}\bm{e_l}) ,
\end{equation}
where $\circ$ denotes an element-wise product and where $\bm{U}$ and $\bm{V}$ are trainable weight matrices. We choose MLB attention because it is parameter efficient and has shown strong empirical performance~\cite{kim2016hadamard,kafle2017visual}.

\subsection{Feature-wise Linear Modulation}
Feature-wise Linear Modulation was introduced in the context of image stylization~\cite{dumoulin2017learned} and extended and shown to be highly effective for multi-modal tasks such as visual question-answering~\cite{devries2017modulating,perez2018film,delbrouck2017modulating}. 

A Feature-wise Linear Modulation (FiLM) layer applies a per-channel scaling and shifting to the convolutional feature maps. Such layers are parameter efficient (only two scalars per feature map) while still retaining high capacity, as they are able to scale up or down, zero-out, or negate whole feature maps. In vision-and-language tasks, another network, the so-called FiLM generator $h$, predicts these modulating parameters from the linguistic input $\bm{e}_l$. More formally, a FiLM layer computes a modulated feature map $\bm{\hat{F}}_{w,h,c}$ as follows:
  \begin{align}
      [\;\bm{\gamma} \;;\; \bm{\beta}\;] = h(\bm{e}_l) \qquad ; \qquad 
    \bm{\hat{F}}_{.,.,c} = \gamma_{c}\bm{F}_{.,.,c} + \beta_{c},
  \end{align}
where $\bm{\gamma}$ and $\bm{\beta}$ are the scaling and shifting parameters which modulate the activations of the original feature map $\bm{F}_{.,.,c}$. We will use the superscript $k \in [1;K]$ to refer to the $k^{th}$ FiLM layer in the network.

FiLM layers may be inserted throughout the hierarchy of a convolutional network, either pre-trained and fixed~\cite{de2017guesswhat} or trained from scratch~\cite{perez2018film}.
Prior FiLM-based models~\cite{devries2017modulating,perez2018film,delbrouck2017modulating} have used a single-hop FiLM generator to predict the FiLM parameters in all layers, e.g.\@, an MLP which takes the language embedding $\bm{e}_{l}$ as input~\cite{devries2017modulating,perez2018film,delbrouck2017modulating}.

\section{\memfilmshort}
\label{sec:mem_cell}

\begin{figure}[t]
\centering
    \includegraphics[width=0.75\linewidth]{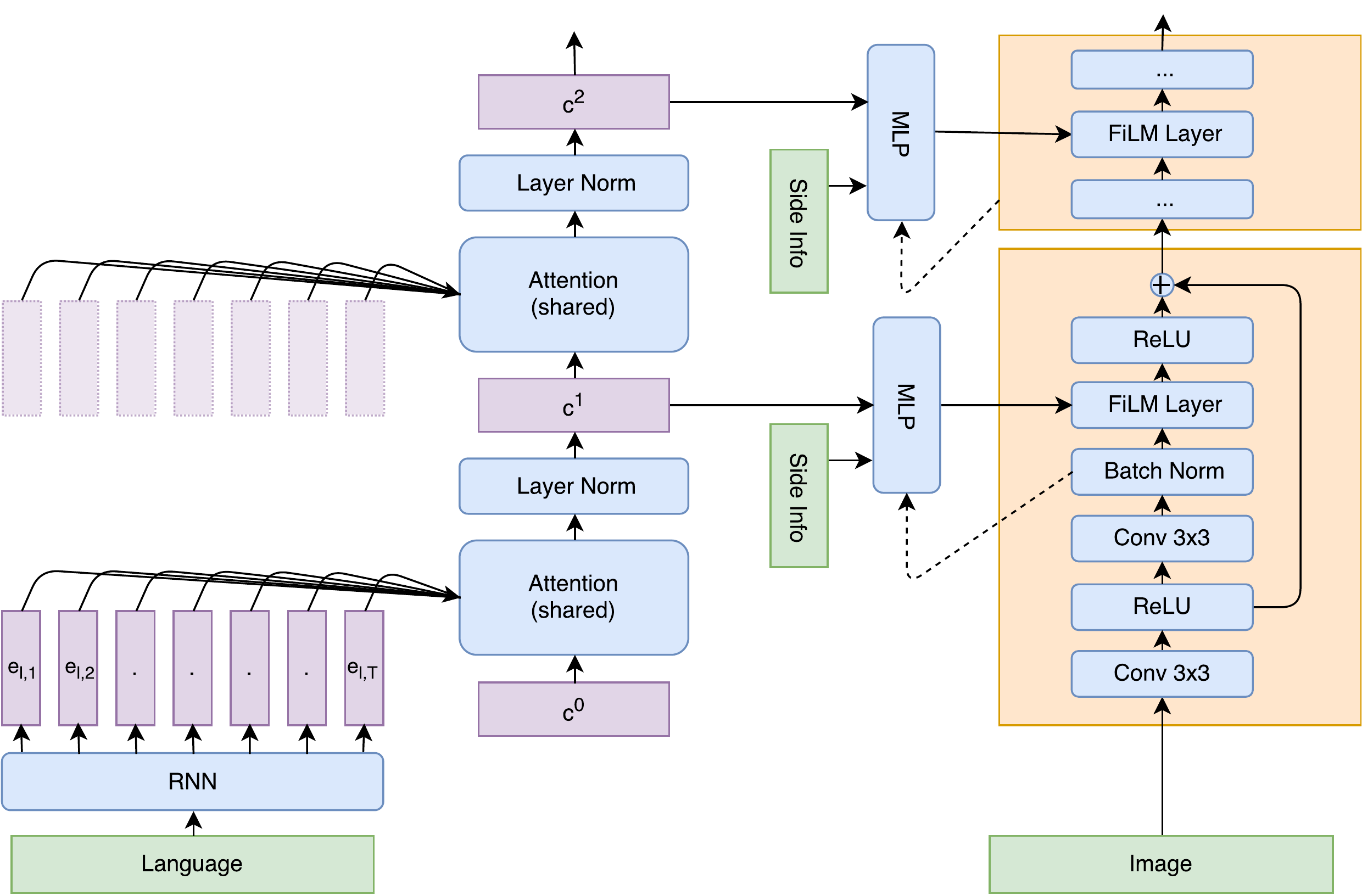}
\caption{\label{fig:film} The \memfilmshort architecture, illustrating inputs (green), layers (blue), and activations (purple). In contrast, \sfilm models predict FiLM parameters directly from $\bm{e}_{l,T}$.}
\end{figure}

In this section, we introduce the \memfilmshort architecture (shown in Fig.~\ref{fig:film}) to predict the parameters of FiLM layers in an iterative fashion, to better scale to longer input sequences such as in dialogue. Another motivation was to better disantangle the linguistic reasoning from the visual one by iteratively attending to both pipelines.


We introduce a context vector $\bm{c}^k$ that acts as a controller for the linguistic and visual pipelines. We initialize the context vector with the final state of a bidirectional RNN $\bm{s}_T$ and repeat the following procedure for each of the FiLM layers in sequence (from lowest to highest convolutional layer): first, the context vector is updated by performing attention over RNN states (extracting relevant language information), and second, the context is used to predict a layer's FiLM parameters (dynamically modulating the visual information). 
Thus, the context vector enables the model to perform multi-hop reasoning over the linguistic pipeline while iteratively modulating the image features. 
More formally, the context vector is computed as follows:
    \begin{align}
    \begin{cases}
     \bm{c}^0 = \bm{s}_T  \\
     \bm{c}^{k} = \sum_{t} \kappa^k_{t}(\bm{c}^{k-1},\bm{s}_{t}) \bm{s}_{t},
  	\end{cases}
    \end{align}
where:

    \begin{equation}
    	\kappa^k_{t}(\bm{c}^{k-1},\bm{s}_{t}) = \frac{\exp(\chi^k_{t})}{\sum_{t} \exp(\chi^k_{t})} \qquad ; \qquad \chi^k_{t}(\bm{c}^{k-1},\bm{s}_{t}) = MLP_{Attn}(g'(\bm{c}^k,\bm{s}_{t})),
    \end{equation}
where the dependence of $\chi^k_{t}$ on $(\bm{c}^{k-1},\bm{s}_{t})$ may be omitted to simplify notation. $MLP_{Attn}$ is a network (shared across layers) which aids in producing attention weights. $g'$ can be any fusion mechanism that facilitates selecting the relevant context to attend to; here we use a simple dot-product following~\cite{luong2015effective}, so $g'(\bm{c}^k,\bm{s}_{t}) =  \bm{c}^k\circ\bm{s}_{t}$ . Finally, FiLM is carried out using a layer-dependent neural network $MLP_{FiLM}^k$:
\begin{align}
    [\;\bm{\gamma}^k \;;\; \bm{\beta}^k\;] = MLP_{FiLM}^k(\bm{c}^k)  \qquad ; \qquad 
    \bm{\hat{F}}^k_{w,h,c} = \gamma^k_{c}\bm{F}^k_{.,.,c} + \beta^k_{c}.
\end{align}
As a regularization, we append a normalization-layer~\cite{ba2016layer} on top of the context vector after each attention step.

\emph{External information.}
Some tasks provide additional information which may be used to further improve the visual modulation. For instance, \GW provides spatial features of the ground truth object to models which must answer questions about that object. Our model incorporates such features by concatenating them to the context vector before generating FiLM parameters.

\emph{Visual feedback.} 
Inspired by the co-attention mechanism~\cite{lu2016hierarchical,zhuang2017parallel}, we also explore incorporating visual feedback into the \memfilmshort architecture. To do so, we first extract the image or crop features $\bm{F}^k$ (immediately before modulation) and apply a global mean-pooling over spatial dimensions. We then concatenate this visual state into the context vector $\bm{c}^k$ before generating the next set of FiLM parameters.

\section{Experiments}
In this section, we first introduce the \Referit and \GW datasets and respective tasks and then describe our overall \memfilmshort architecture\footnote{The code and hyperparameters are available at \url{https://github.com/GuessWhatGame}}.

\subsection{Dataset}

\emph{\Referit}~\cite{kazemzadeh2014referitgame,yu2016modeling} is a cooperative two-player game. The first player (the Oracle) selects an object in a rich visual scene, for which they must generate an expression that refers to it (e.g.\@, ``the person eating ice cream''). Based on this expression, the second player (the Guesser) must then select an object within the image. There are four \Referit datasets exist: RefClef, RefCOCO, RefCOCO+ and RefCOCOg. The first dataset contains 130K references over 20K images from the ImageClef dataset~\cite{Mller:2012:IEE:2462702}, while the three other datasets respectively contain 142K, 142K and 86K references over 20K, 20k and 27K images from the MSCOCO dataset~\cite{lin2014microsoft}. Each dataset has small differences. RefCOCO and RefClef were constructed using different image sets. RefCOCO+ forbids certain words to prevent object references from being too simplistic, and RefCOCOg only relies on images containing 2-4 objects from the same category. RefCOCOg also contains longer and more complex sentences than RefCOCO (8.4 vs. 3.5 average words). Here, we will show results on both the Guesser and Oracle tasks.

\emph{\GW}~\cite{de2017guesswhat} is a cooperative three-agent game in which players see the picture of a rich visual scene with several objects. One player (the Oracle) is randomly assigned an object in the scene. The second player (Questioner) aims to ask a series of yes-no questions to the Oracle to collect enough evidence to allow the third player (Guesser) to correctly locate the object in the image. The \GW dataset is composed of 131K successful natural language dialogues containing 650k question-answer pairs on over 63K images from MSCOCO~\cite{lin2014microsoft}. Dialogues contain 5.2 question-answer pairs and 34.4 words on average. Here, we will focus on the Guesser and Oracle tasks.

\begin{figure}[t]
\centering
\includegraphics[width=0.85\linewidth]{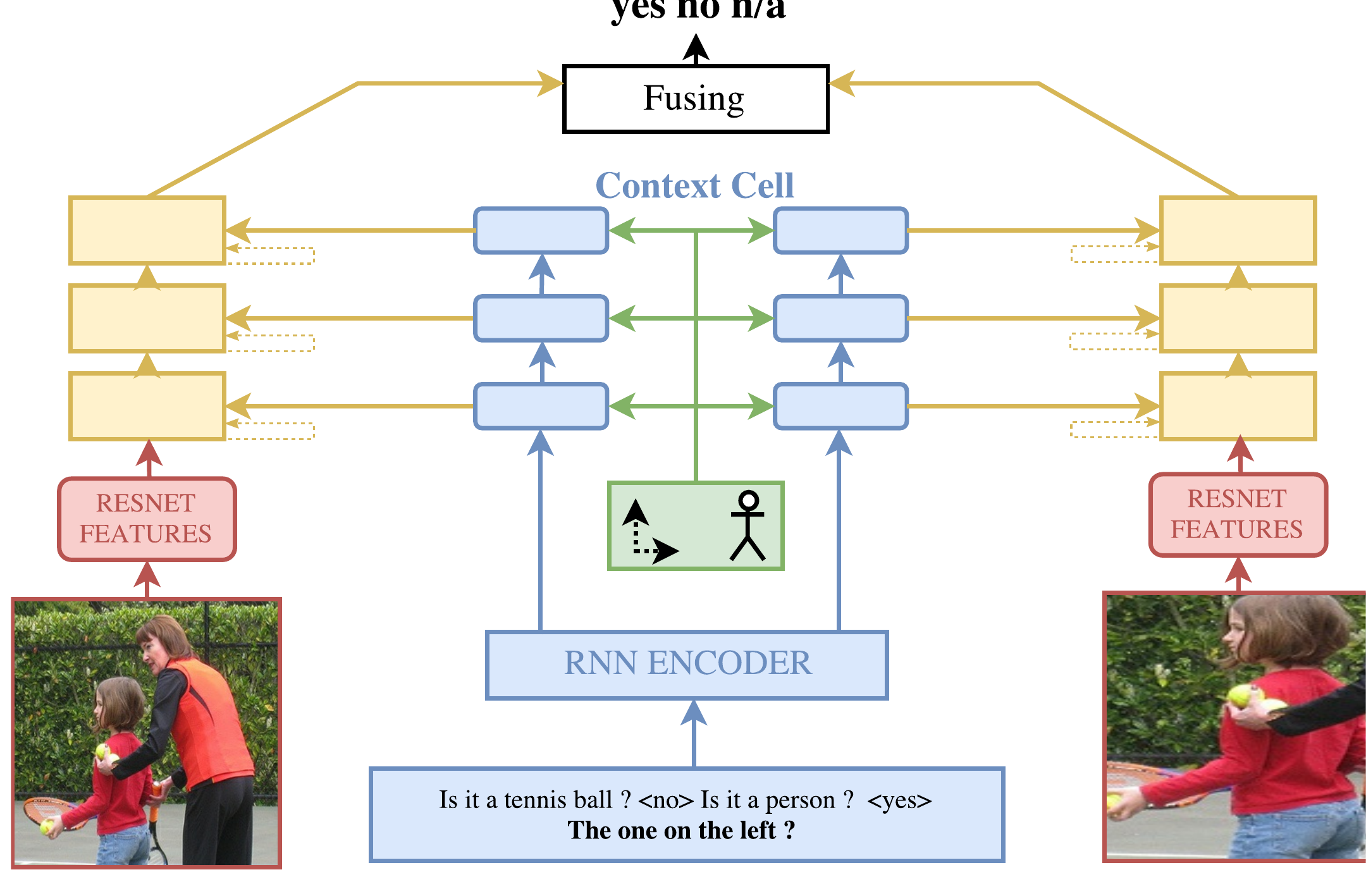}
\caption{Overall model, consisting of a visual pipeline (red and yellow) and linguistic pipeline (blue) and incorporating additional contextual information (green).}
\label{fig:oracle model}
\end{figure}

\subsection{Task Descriptions}

\emph{Game Features.} Both games consist of triplets $(\Image, l, o)$, where $\Image \in \mathbb{R}^{3\times M\times N}$ is an RGB image and $l$ is some language input (i.e.\@, a series of words) describing an object $o$ in $\Image$. 
The object $o$ is defined by an object category, a pixel-wise segmentation, an RGB crop of $\Image$ based on bounding box information, and hand-crafted spatial information $\bm{x}_{spatial}$, where
\begin{equation}
\bm{x}_{spatial} = [ x_{min}, y_{min}, x_{max}, y_{max}, x_{center}, y_{center}, w_{box}, h_{box}]
\end{equation}
We replace words with two or fewer occurrences with an ${<}unk{>}$ token.

\emph{The Oracle task.} Given an image $\Image$, an object $o$, a question $q$, and a sequence $\delta$ of previous question-answer pairs $(\bm{q},a)_{{1:\delta}}$ where $a \in \{\mbox{Yes},\mbox{No},\mbox{N/A}\}$, the oracle's task is to produce an answer $a$ that correctly answers the question $q$. 

\emph{The Guesser task.} Given an image $\Image$, a list of objects $O = o_{1:\Phi}$, a target object $o^* \in O$ and the dialogue $\mathcal{D}$, the guesser needs to output a probability $\sigma_\phi$ that each object $o_\phi$ is the target object $o^*$. Following~\cite{hu2016natural}, the Guesser is evaluated by selecting the object with the highest probability of being correct. Note that even if the individual probabilities $\sigma_\phi$ are between 0 and 1, their sum can be greater than 1. More formally, the Guesser loss and error are computed as follows:
\begin{equation}
L_{Guesser} = \frac{-1}{N_{games}}\sum_n^{N_{games}} \frac{1}{\Phi^n} \sum_\phi^\Phi \text{log}(p(o^*|\Image^n, o^n_\phi, \mathcal{D}^n)) 
\end{equation}
\begin{equation}
E_{Guesser} = \frac{-1}{N_{games}}\sum_n^{N_{games}} \mathbb{1}( o^* \not = o_{ \argmax_\phi \sigma^n_\phi})
\end{equation}
where $\mathbb{1}$ is the indicator function and $\Phi^n$ the number of objects in the $n^{th}$ game.

\subsection{Model}
We use similar models for both \Referit and \GW and provide its architectural details in this subsection.

\emph{Object embedding} The object category is fed into a dense look-up table $\bm{e}_{cat}$, and the spatial information is scaled to [-1;1] before being up-sampled via non-linear projection to $\bm{e}_{spat}$. We do not use the object category in \Referit models.

\emph{Visual Pipeline} We first resized the image and object crop to $448\times 448$ before extracting $14\times 14\times 1024$ dimensional features from a ResNet-152~\cite{he2016deep} (block3) pre-trained on ImageNet~\cite{russakovsky2015imagenet}. Following~\cite{perez2018film}, we feed these features to a $3\times3$ convolution layer with Batch Normalization~\cite{ioffe2015batch} and Rectified Linear Unit~\cite{nair2010rectified} (ReLU). We then stack four modulated residual blocks (shown in Fig~\ref{fig:film}), each producing a set of feature maps $\bm{F}^k$ via (in order) a $1\times 1$ convolutional layer (128 units), ReLU activations, a $3\times 3$ convolutional layer (128 units), and an untrainable Batch Normalization layer. The residual block then modulates $\bm{F}^k$ with a FiLM layer to get $\bm{\hat{F}}^k$, before again applying ReLU activations. Lastly, a residual connection sums the activations of both ReLU outputs. After the last residual block, we use a $1\times1$ convolution layer (512 units) with Batch Normalization and ReLU followed by MLB attention~\cite{kim2016hadamard} (256 units and 1 glimpse) to obtain the final embedding $\bm{e}_v$. Note our model uses two independent visual pipeline modules: one to extract modulated image features $\bm{e}^{img}_v$, one to extract modulated crop features $\bm{e}^{crop}_v$.

To incorporate spatial information, we concatenate two coordinate feature maps indicating relative x and y spatial position (scaled to $[-1, 1]$) with the image features before each convolution layer (except for convolutional layers followed by FiLM layers). In addition, the pixel-wise segmentations $S \in \{0,1\}^{M\times N}$ are rescaled to $14\times 14$ floating point masks before being concatenated to the feature maps.

\emph{Linguistic Pipeline} We compute the language embedding by using a word-embedding look-up (200 dimensions) with dropout followed by a Bi-GRU ($512\times2$units) with Layer Normalization~\cite{ba2016layer}. As described in Section~\ref{sec:mem_cell}, we initialize the context vector with the last RNN state $\bm{c}^0=\bm{s}_{T}$. We then attend to the other Bi-GRU states via an attention mechanism with a linear projection and ReLU activations and regularize the new context vector with Layer Normalization.

\emph{FiLM parameter generation} We concatenate spatial information $\bm{e}_{spat}$ and object category information $\bm{e}_{cat}$ to the context vector. In some experiments, we also concatenate a fourth embedding consisting of intermediate visual features $\bm{F}^k$ after mean-pooling. Finally, we use a linear projection to map the embedding to FiLM parameters.

\emph{Final Layers} We first generate our final embedding by concatenating the output of the visual pipelines $\bm{e}_{final} = [\bm{e}^{img}_v$ ; $\bm{e}^{crop}_v]$ before applying a linear projection (512 units) with ReLU and a softmax layer. 

\emph{Training Process} We train our model end-to-end with Adam~\cite{kingma2014adam} (learning rate $3e^{-4}$), dropout ($0.5$), weight decay ($5e^{-6}$) for convolutional network layers, and a batch size of 64. We report results after early stopping on the validation set with a maximum of 15 epochs. 

\subsection{Baselines}
In our experiments, we re-implement several baseline models to benchmark the performance of our models.
The standard \emph{Baseline NN} simply concatenates the image and object crop features after mean pooling, the linguistic embedding, and the spatial embedding and the category embedding (\GW only), passing those features to the same final layers described in our proposed model. We refer to a model which uses the MLB attention mechanism to pool the visual features as \emph{Baseline NN+MLB}.
We also implement a \emph{\sfilm} mechanism which is equivalent to setting all context vectors equal to the last state of the Bi-GRU $\bm{e}_{l,T}$. Finally, we experiment with injecting intermediate visual features into the FiLM Generator input, and we refer to the model as \emph{\memfilmshort~(+img)}.



\begin{table}[t]
      \caption{\Referit Guesser Error.}
      \label{tab:referit_res}
      \centering
       \scriptsize
\begin{tabular}{l|x{1.2cm}x{1.1cm}x{1.1cm}|x{1.1cm}x{1.1cm}x{1.1cm}|x{1.5cm}}
        \toprule
        \textbf{Referit} & \multicolumn{3}{c|}{RefCOCO} & \multicolumn{3}{c|}{RefCOCO+} &\multicolumn{1}{c}{RefCOCOg}\\
        Split by & \multicolumn{3}{c|}{(unc)} & \multicolumn{3}{c|}{(unc)} &\multicolumn{1}{c}{(google)} \\
        Report on& Valid& TestA & TestB & Valid & TestA & TestB & Val  \\
        \midrule
        MMI~\cite{nagaraja2016modeling}          &  -     &71.7\% & 71.1\% &  -      & 58.4\% & 51.2\% & 59.3\% \\ 
        visDif + MMI~\cite{yu2016modeling}       &  -     &74.6\% & 76.6\% &  -      & 59.2\% & 55.6\% & 64.0\% \\
        NEG Bag~\cite{nagaraja2016modeling}      &  -     &75.6\% & 78.0\% &  -      & -      &  -     & 68.4\% \\
        Joint-SLR~\cite{yu2017joint}             & 78.9\% &78.0\% & 80.7\% &  61.9\% & 64.0\% & 59.2\% & - \\
        PLAN~\cite{zhuang2017parallel}           & 81.7\% &80.8\% & 81.3\% &  64.2\% & 66.3\% & 61.5\% & 69.5\% \\
        MAttN~\cite{yu2018mattnet}               & $\bm{85.7}$\% &85.3\% & $\bm{84.6}$\% &  71.0\% & 75.1\% & $\bm{66.2}$\% & - \\
        \midrule  
        Baseline NN+MLB                         & 77.6\% &79.6\%&77.2\%         &60.8\%       & 59.7\%   &66.2\%   & 63.1\%    \\
        \sfilm                                   & 83.4\% &85.8\% &80.9\%         &72.1\%       & 77.3\%   &63.9\%   & 67.8\%    \\
        \memfilmshort                            & 83.5\% &86.5\% &81.3\%        &73.4\%       & 77.7\%   &64.5\%   & 69.8\%    \\
        \memfilmshort (+img)                     & 84.9\% &$\bm{87.4}$\% &83.1\% &$\bm{73.8\%}$&$\bm{78.7}$\%    &65.8\%   & $\bm{71.5}$\%    \\
        \bottomrule
        \end{tabular}
\end{table}

\subsection{Results}

\emph{\Referit Guesser} We report the best test error of the outlined methods on the \Referit Guesser task in Tab.~\ref{tab:referit_res}. Note that RefCOCO and RefCOCO+ split test sets into TestA and TestB, only including expression referring towards people and objects, respectively. We do not report~\cite{yu2018mattnet} and~\cite{yu2017joint} scores on RefCOCOg as the authors use a different split (umd). Our initial baseline achieves 77.6\%, 60.8\%, 63.1\%, 73.4\% on the RefCOCO, RefCOCO+, RefCOCOg, RefClef datasets, respectively, performing comparably to state-of-the-art models. We observe a significant improvements using a FiLM-based architecture, jumping to 84.9\%, 87.4\%, 73.8\%, 71.5\%, respectively, and outperforming most prior methods and achieving comparably performance with the concurrent MAttN~\cite{yu2018mattnet} model. Interestingly, MAttN and \memfilmshort are built in two different manners; while the former has three specialized reasoning blocks, our model uses a generic feature modulation approach. These architectural differences surface when examining test splits: MAttN achieves excellent results on referring expression towards objects while \memfilmshort performs better on referring expressions towards people.  

\begin{table*}[t]
    \caption{\label{tab:oracle_results} \GW Oracle Error by Model and Input Type.}
        \centering
        \scriptsize
		\begin{tabular}{l|x{1.1cm}x{1.1cm}x{1.1cm}x{1.1cm}x{1.1cm} | x{ 1.5cm} }
        \toprule
        \textbf{Oracle Models} & Quest. & Dial.  & Object  & Image   & Crop   & Test Error \\
        \midrule
        Dominant class (``no'')   & \xmark & \xmark & \xmark   & \xmark  & \xmark & 50.9\% \\
        Question only~\cite{de2017guesswhat} & \cmark & \xmark & \xmark   & \xmark  & \xmark & 41.2\% \\
        Image only~\cite{de2017guesswhat}    & \xmark & \xmark & \xmark   & \cmark  & \xmark & 46.7\% \\
        Crop only~\cite{de2017guesswhat}     & \xmark & \xmark & \xmark   & \xmark  & \cmark & 43.0\% \\
        \midrule
        No-Vision (Quest.)~\cite{de2017guesswhat} & \cmark & \xmark & \cmark   & \xmark  & \xmark &21.5\% \\
        No-Vision (Dial.)                        & \xmark & \cmark & \cmark   & \xmark  & \xmark &20.6\% \\
        \midrule
        Baseline NN (Quest.)         & \cmark & \xmark & \cmark   & \cmark  & \cmark & 23.3\%\\
        Baseline NN (Dial.)          & \xmark & \cmark & \cmark   & \cmark  & \cmark & 22.4\%\\
        Baseline NN + MLB (Quest.)   & \cmark & \xmark & \cmark   & \cmark  & \cmark & 21.8\%\\
        Baseline NN + MLB (Dial.)    & \xmark & \cmark & \cmark   & \cmark  & \cmark & 21.1\%\\
        \midrule
        MODERN~\cite{devries2017modulating}  & \cmark & \xmark & \cmark   & \xmark  & \cmark &19.5\% \\
        \midrule
        \sfilm (Quest.)			      & \cmark & \xmark & \cmark   & \cmark  & \cmark & 17.8\%\\
        \sfilm (Dial.)			      & \xmark & \cmark & \cmark   & \cmark  & \cmark & 17.6\%\\
        \memfilmshort 		      & \xmark & \cmark & \cmark   & \cmark  & \cmark & \bf{16.9\%} \\
        \memfilmshort (+img) & \xmark  & \cmark & \cmark  & \cmark   & \cmark & 17.1\% \\
        \bottomrule     
        \end{tabular}
\end{table*}

\emph{\GW Oracle} We report the best test error of several variants of \GW Oracle models in Tab.~\ref{tab:oracle_results}. First, we baseline any visual or language biases by predicting the Oracle's target answer using only the image (46.7\% error) or the question (41.1\% error). As first reported in~\cite{de2017guesswhat}, we observe that the baseline methods perform worse when integrating the image and crop inputs (21.1\%) rather than solely using the object category and spatial location (20.6\%). On the other hand, concatenating previous question-answer pairs to answer the current question is beneficial in our experiments. Finally, using \sfilm reduces the error to 17.6\% and \memfilmshort further to 16.9\%, outperforming the previous best model by 2.4\%.

\emph{\GW Guesser} We provide the best test error of the outlined methods on the \GW Guesser task in Tab.~\ref{tab:guesser_results}. As a baseline, we find that random object selection achieves an error rate of 82.9\%. Our initial model baseline performs significantly worse (38.3\%) than concurrent models (36.6\%), highlighting that successfully jointly integrating crop and image features is far from trivial. However, \sfilm manages to lower the error to 35.6\%. Finally, \memfilmshort architecture  outperforms other models with a final error of 30.5\%.

\section{Discussion}

\emph{\sfilm \textit{vs.} \memfilmshort} In the \GW task, \memfilmshort outperforms \sfilm by 6.1\% on the Guesser task but only 0.7\% on the Oracle task. We think that the small performance gain for the Oracle task is due to the nature of the task; to answer the current question, it is often not necessary to look at previous question-answer pairs, and in most cases this task does not require a long chain of reasoning. On the other hand, the Guesser task needs to gather information across the whole dialogue in order to correctly retrieve the object, and it is therefore more likely to benefit from multi-hop reasoning. The same trend can be observed for \Referit. \sfilm and \memfilmshort perform similarly on RefClef and RefCOCO, while we observe 1.3\% and 2\% gains on RefCOCO+ and RefCOCOg, respectively. This pattern of performance is intuitive, as the former datasets consist of shorter referring expressions (3.5 average words) than the latter (8.4 average words in RefCOCOg), and the latter datasets also consist of richer, more complex referring expressions due e.g.\@ to taboo words (RefCOCO+).
In short, our experiments demonstrate that \memfilmshort is better able reason over complex linguistic sequences.

\begin{table*}[t]
 \caption{\GW Guesser Error.}
 \label{tab:guesser_results}
    \begin{minipage}{0.36\linewidth}
        \centering
        \scriptsize
		\begin{tabular}{l|x{1.5cm}}
        \toprule
        \textbf{Guesser Error} & {Test Error}  \\
        \midrule
        Random & 82.9\% \\
        \midrule
        LSTM~\cite{de2017guesswhat}  &38.7\% \\
        LSTM + Img~\cite{de2017guesswhat}  &39.5\% \\
        PLAN~\cite{zhuang2017parallel}  &36.6\% \\
        \midrule
        Base NN + MLB (crop)& 38.3\%\\
        \sfilm & 35.6\%\\
        \memfilmshort &\bf{30.5\%} \\
        \bottomrule     
        \end{tabular}   
    \end{minipage}%
    \quad
    \begin{minipage}{0.62\linewidth}
       \centering
       \scriptsize
\begin{tabular}{l|x{1.2cm}x{1.2cm}x{1.45cm}}
        \toprule
        \textbf{Guesser Error} & {Crop} & {Image} & {Crop+Img} \\
        \midrule
        Baseline NN                 & 38.3\% & 40.0\% & 45.1\%        \\
        \sfilm                   & 35.3\% & 35.7\% & 35.6\%        \\
        \memfilmshort            & 32.3\% & 35.0\% & $\bf{30.5\%}$ \\
        \memfilmshort (no categ.)   & 33.1\% & 40\% & 33.4\%\\
        \bottomrule
        \end{tabular}        

    \end{minipage}%
    \end{table*}

\begin{figure}[t]
\centering
    \subfloat{\includegraphics[width=0.44\linewidth]{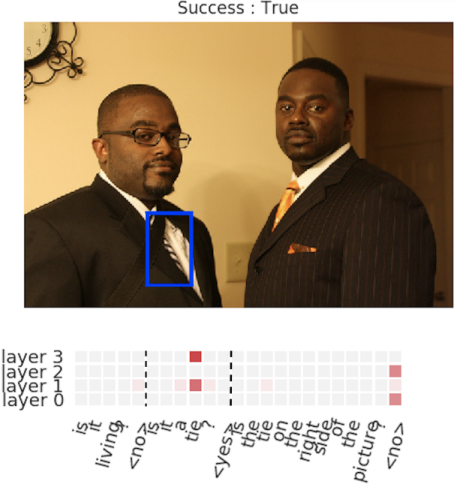}}
        \quad
    \subfloat{\includegraphics[width=0.45\linewidth]{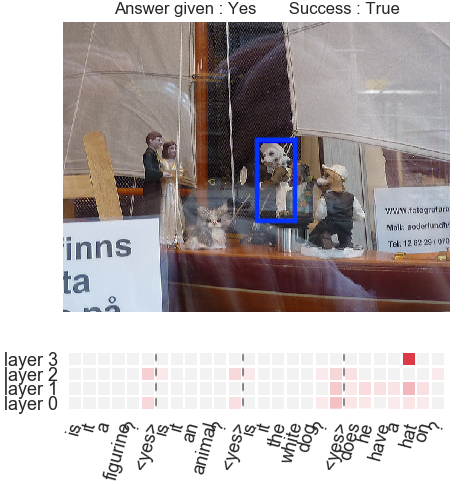}}
\caption{\label{fig:attention} Guesser (left) and Oracle (right) attention visualizations for the visual pipeline which processes the object crop.}
\end{figure}

\emph{Reasoning mechanism} 
We conduct several experiments to better understand our method. First, we assess whether \memfilmshort performs better because of increased network capacity. We remove the attention mechanism over the linguistic sequence and update the context vector via a shared MLP. We observe that this change significantly hurts performance across all tasks, e.g.\@, increasing the \memfilmshort error of the Guesser from 30.5 to 37.3\%.
Second, we investigate how the model attends to \GW dialogues for the Oracle and Guesser tasks, providing more insight into how to the model reasons over the language input. We first look at the top activation in the (crop) attention layers to observe where the most prominent information is. Note that similar trends are observed for the image pipeline. As one would expect, the Oracle is focused on a specific word in the last question 99.5\% of the time, one which is crucial to answer the question at hand. However, this ratio drops to 65\% in the Guesser task, suggesting the model is reasoning in a different way. If we then extract the top 3 activations per layer, the attention points to ${<}yes{>}$ or ${<}no{>}$ tokens (respectively) at least once, 50\% of the time for the Oracle and Guesser, showing that the attention is able to correctly split the dialogue into question-answer pairs. Finally, we plot the attention masks for each FiLM layer to have a better intuition of this reasoning process in Fig.~\ref{fig:attention}.

\emph{Crop \textit{vs.} Image.} We also evaluate the impact of using the image and/or crop on the final error for the Guesser task~\ref{tab:guesser_results}. Using the image alone (while still including object category and spatial information) performs worse than using the crop. However, using image and crop together inarguably gives the lowest errors, though prior work has not always used the crop due to architecture-specific GPU limitations~\cite{devries2017modulating}.

\emph{Visual feedback} We explore whether adding visual feedback to the context embedding improves performance. While it has little effect on the \GW Oracle and Guesser tasks, it improves the accuracy on \Referit by 1-2\%. Note that \Referit does not include class labels of the selected object, so the visual feedback might act as a surrogate for this information. To further investigate this hypothesis, we remove the object category from the \GW task and report results in Tab.~\ref{tab:oracle_results_nocat} in the supplementary material. In this setup, we indeed observe a relative improvement 0.4\% on the Oracle task, further confirming this hypothesis. 


\emph{Pointing Task} In \GW, the Guesser must select an object from among a list of items. A more natural task would be to have the Guesser directly point out the object as a human might. Thus, in the supplementary material, we introduce this task and provide initial baselines (Tab.~\ref{sec:pointing}) which include FiLM models. This task shows ample room for improvement with a best test error of 84.0\%.

\section{Related Work}

The \Referit game~\cite{kazemzadeh2014referitgame} has been a testbed for various vision-and-language tasks over the past years, including object retrieval~\cite{nagaraja2016modeling,yu2016modeling,yu2017joint,zhuang2017parallel,luo2017comprehension,yu2018mattnet}, semantic image segmentation~\cite{hu2016segmentation,rohrbach2016grounding}, and generating referring descriptions~\cite{yu2016modeling,luo2017comprehension,yu2017joint}.
To tackle object retrieval,~\cite{nagaraja2016modeling,yu2016modeling,yu2018mattnet} extract additional visual features such as relative object locations and~\cite{yu2017joint,luo2017comprehension} use reinforcement learning to iteratively train the object retrieval and description generation models.
Closer to our work,~\cite{hu2016natural,zhuang2017parallel} use the full image and the object crop to locate the correct object. While some previous work relies on task-specific modules~\cite{yu2016modeling,yu2018mattnet}, our approach is general and can be easily extended to other vision-and-language tasks.

The \GW game~\cite{de2017guesswhat} can be seen as a dialogue version of the \Referit game, one which additionally draws on visual question answering ability.
\cite{strub2017end,lee2018answerer,zhu2017interactive} make headway on the dialogue generation task via reinforcement learning. However, these approaches are
bottlenecked by the accuracy of Oracle and Guesser models, despite existing modeling advances~\cite{zhuang2017parallel,devries2017modulating}; accurate Oracle and Guesser models are crucial for providing a meaningful learning signal for dialogue generation models, so we believe the \memfilmshort architecture will facilitate high quality dialogue generation as well.

A special case of Feature-wise Linear Modulation was first successfully applied to image style transfer~\cite{dumoulin2017learned}, whose approach modulates image features according to some image style (i.e.\@, cubism or impressionism).
\cite{devries2017modulating} extended this approach to vision-and-language tasks, injecting FiLM-like layers along the entire visual pipeline of a pre-trained ResNet.
\cite{perez2018film} demonstrates that a convolutional network with FiLM layers achieves strong performance on CLEVR~\cite{johnson2017clevr}, a task that focuses on answering reasoning-oriented, multi-step questions about synthetic images.
Subsequent work has demonstrated that FiLM and variants thereof are effective for video object segmentation where the conditioning input is the first image's segmentation (instead of language)~\cite{yang2018osmn} and language-guided image segmentation~\cite{rupprecht2018guide}. Even more broadly,~\cite{dumoulin2018feature-wise} overviews the strength of FiLM-related methods across machine learning domains, ranging from reinforcement learning to generative modeling to domain adaptation.

There are other notable models that decompose reasoning into different modules. For instance, Neural Turing Machines~\cite{graves2014neural,graves2016hybrid} divide a model into a controller with read and write units. Memory networks use an attention mechanism to answer a query by reasoning over a linguistic knowledge base~\cite{weston2014memory,sukhbaatar2015end} or image features~\cite{xiong2016dynamic}. A memory network updates a query vector by performing several attention hops over the memory before outputting a final answer from this query vector. Although \memfilmshort computes a similar context vector, this intermediate embedding is used to predict FiLM parameters rather than the final answer. Thus, \memfilmshort includes a second reasoning step over the image.

Closer to our work, \cite{arad2018compositional} designed networks composed of Memory, Attention, and Control (MAC) cells to perform visual reasoning. Similar to Neural Turing Machines, each MAC cell is composed of a control unit that attends over the language input, a read unit that attends over the image and a write unit that fuses both pipelines. Though conceptually similar to \memfilmshort models, Compositional Attention Networks differ structurally, for instance using a dynamic neural architecture and relying on spatial attention rather than FiLM.

\section{Conclusion}
In this paper, we introduce a new way to exploit Feature-wise Linear Modulation (FiLM) layers for vision-and-language tasks. Our approach generates the parameters of FiLM layers going up the visual pipeline by attending to the language input in multiple hops rather than all at once.
We show \memfilm architectures are better able to handle longer sequences than their single-hop counterparts. We outperform state-of-the-art vision-and-language models significantly on the long input sequence \GW tasks, while maintaining state-of-the-art performance for the shorter input sequence \Referit task. Finally, this \memfilm approach uses few problem-specific priors, and thus we believe it can extended to a variety of vision-and-language tasks, particularly those requiring complex visual reasoning.

\paragraph*{Acknowledgements}
The authors would like to acknowledge the stimulating research environment of the SequeL Team. We also thank Vincent Dumoulin for helpful discussions. 
We acknowledge the following agencies for research funding and computing support: Project BabyRobot (H2020-ICT-24-2015, grant agreement no.687831), CHISTERA IGLU and CPER Nord-Pas de Calais/FEDER DATA Advanced data science and technologies 2015-2020, NSERC, Calcul Qu\'{e}bec, Compute Canada, the Canada Research Chairs, and CIFAR.

\bibliographystyle{splncs04}
\bibliography{egbib}

\begin{thebibliography}{10}
\providecommand{\url}[1]{\texttt{#1}}
\providecommand{\urlprefix}{URL }
\providecommand{\doi}[1]{https://doi.org/#1}

\bibitem{antol2015vqa}
Antol, S., Agrawal, A., Lu, J., Mitchell, M., Batra, D., Lawrence~Zitnick, C.,
  Parikh, D.: Vqa: Visual question answering. In: Proc. of ICCV (2015)

\bibitem{ba2016layer}
Ba, J.L., Kiros, J.R., Hinton, G.E.: Layer normalization. Deep Learning
  Symposium (NIPS)  (2016)

\bibitem{bahdanau2014neural}
Bahdanau, D., Cho, K., Bengio, Y.: Neural machine translation by jointly
  learning to align and translate. In: Proc. of ICLR (2015)

\bibitem{chung2014empirical}
Chung, J., Gulcehre, C., Cho, K., Bengio, Y.: Empirical evaluation of gated
  recurrent neural networks on sequence modeling. In: Proc. of ICML (2015)

\bibitem{das2017visual}
Das, A., Kottur, S., Gupta, K., Singh, A., Yadav, D., Moura, J.M., Parikh, D.,
  Batra, D.: Visual dialog. In: Proc. of CVPR (2017)

\bibitem{de2017guesswhat}
De~Vries, H., Strub, F., Chandar, S., Pietquin, O., Larochelle, H., Courville,
  A.: Guesswhat?! visual object discovery through multi-modal dialogue. In:
  Proc. of CVPR (2017)

\bibitem{delbrouck2017modulating}
Delbrouck, J.B., Dupont, S.: Modulating and attending the source image during
  encoding improves multimodal translation. Visually-Grounded Interaction and
  Language Workshop (NIPS)  (2017)

\bibitem{dumoulin2017learned}
Dumoulin, V., Shlens, J., Kudlur, M.: {A Learned Representation For Artistic
  Style}. In: Proc. of ICLR (2017)

\bibitem{dumoulin2018feature-wise}
Dumoulin, V., Perez, E., Schucher, N., Strub, F., Vries, H.d., Courville, A.,
  Bengio, Y.: Feature-wise transformations. Distill  (2018).
  \doi{10.23915/distill.00011},
  https://distill.pub/2018/feature-wise-transformations

\bibitem{everingham2010pascal}
Everingham, M., Van~Gool, L., Williams, C.K., Winn, J., Zisserman, A.: The
  pascal visual object classes (voc) challenge. International journal of
  computer vision  \textbf{88}(2),  303--338 (2010)

\bibitem{fukui2016multimodal}
Fukui, A., Park, D.H., Yang, D., Rohrbach, A., Darrell, T., Rohrbach, M.:
  Multimodal compact bilinear pooling for visual question answering and visual
  grounding. In: Proc. of EMNLP (2016)

\bibitem{girshick2014rich}
Girshick, R., Donahue, J., Darrell, T., Malik, J.: Rich feature hierarchies for
  accurate object detection and semantic segmentation. In: Proc. of of CVPR
  (2014)

\bibitem{graves2014neural}
Graves, A., Wayne, G., Danihelka, I.: Neural turing machines. arXiv preprint
  arXiv:1410.5401  (2014)

\bibitem{graves2016hybrid}
Graves, A., Wayne, G., Reynolds, M., Harley, T., Danihelka, I.,
  Grabska-Barwi{\'n}ska, A., Colmenarejo, S.G., Grefenstette, E., Ramalho, T.,
  Agapiou, J., et~al.: Hybrid computing using a neural network with dynamic
  external memory. Nature  \textbf{538}(7626), ~471 (2016)

\bibitem{he2016deep}
He, K., Zhang, X., Ren, S., Sun, J.: Deep residual learning for image
  recognition. In: Proc. of CVPR (2016)

\bibitem{hu2016segmentation}
Hu, R., Rohrbach, M., Darrell, T.: Segmentation from natural language
  expressions. In: Proc. of ECCV (2016)

\bibitem{hu2016natural}
Hu, R., Xu, H., Rohrbach, M., Feng, J., Saenko, K., Darrell, T.: Natural
  language object retrieval. In: Proc. of CVPR (2016)

\bibitem{arad2018compositional}
Hudson, D.A., Manning, C.D.: Compositional attention networks for machine
  reasoning. In: Proc. of ICL (2018)

\bibitem{ioffe2015batch}
Ioffe, S., Szegedy, C.: Batch normalization: Accelerating deep network training
  by reducing internal covariate shift. In: Proc. of ICML (2015)

\bibitem{jabri16vqa}
Jabri, A., Joulin, A., van~der Maaten, L.: Revisiting visual question answering
  baselines. In: Proc. of ECCV (2016)

\bibitem{johnson2017clevr}
Johnson, J., Hariharan, B., van~der Maaten, L., Fei-Fei, L., Zitnick, C.L.,
  Girshick, R.: Clevr: A diagnostic dataset for compositional language and
  elementary visual reasoning. In: Proc. of CVPR (2017)

\bibitem{kafle2017visual}
Kafle, K., Kanan, C.: Visual question answering: Datasets, algorithms, and
  future challenges. Computer Vision and Image Understanding  \textbf{163},
  3--20 (2017)

\bibitem{kazemzadeh2014referitgame}
Kazemzadeh, S., Ordonez, V., Matten, M., Berg, T.: Referitgame: Referring to
  objects in photographs of natural scenes. In: Proc. of EMNLP (2014)

\bibitem{kim2016hadamard}
Kim, J.H., On, K.W., Lim, W., Kim, J., Ha, J.W., Zhang, B.T.: {Hadamard Product
  for Low-rank Bilinear Pooling}. In: Proc. of ICLR (2017)

\bibitem{kim2016multimodal}
Kim, J.H., Lee, S.W., Kwak, D., Heo, M.O., Kim, J., Ha, J.W., Zhang, B.T.:
  Multimodal residual learning for visual qa. In: Proc. of NIPS (2016)

\bibitem{kingma2014adam}
Kingma, D.P., Ba, J.: Adam: A method for stochastic optimization. In: Proc. of
  ICLR (2014)

\bibitem{krizhevsky2012imagenet}
Krizhevsky, A., Sutskever, I., Hinton, G.E.: Imagenet classification with deep
  convolutional neural networks. In: Proc. of of NIPS (2012)

\bibitem{lee2018answerer}
Lee, S.W., Heo, Y.J., Zhang, B.T.: Answerer in questioner's mind for
  goal-oriented visual dialogue. Visually-Grounded Interaction and Language
  Workshop (NIPS)  (2018)

\bibitem{lin2014microsoft}
Lin, T.Y., Maire, M., Belongie, S., Hays, J., Perona, P., Ramanan, D.,
  Doll{\'a}r, P., Zitnick, C.L.: Microsoft coco: Common objects in context. In:
  Proc. of ECCV (2014)

\bibitem{long2015fully}
Long, J., Shelhamer, E., Darrell, T.: Fully convolutional networks for semantic
  segmentation. In: Proc. of CVPR (2015)

\bibitem{lu2016hierarchical}
Lu, J., Yang, J., Batra, D., Parikh, D.: Hierarchical question-image
  co-attention for visual question answering. In: Proc. of NIPS (2016)

\bibitem{luo2017comprehension}
Luo, R., Shakhnarovich, G.: Comprehension-guided referring expressions. In:
  Proc. of CVPR (2017)

\bibitem{luong2015effective}
Luong, M.T., Pham, H., Manning, C.D.: Effective approaches to attention-based
  neural machine translation. In: Proc. of EMNLP (2015)

\bibitem{malinowski2015ask}
Malinowski, M., Rohrbach, M., Fritz, M.: Ask your neurons: A neural-based
  approach to answering questions about images. In: Proc. of ICCV (2015)

\bibitem{Mller:2012:IEE:2462702}
Mller, H., Clough, P., Deselaers, T., Caputo, B.: ImageCLEF: Experimental
  Evaluation in Visual Information Retrieval. Springer (2012)

\bibitem{nagaraja2016modeling}
Nagaraja, V.K., Morariu, V.I., Davis, L.S.: Modeling context between objects
  for referring expression understanding. In: Proc. of ECCV (2016)

\bibitem{nair2010rectified}
Nair, V., Hinton, G.E.: Rectified linear units improve restricted boltzmann
  machines. In: Proc. of ICML (2010)

\bibitem{perez2018film}
Perez, E., Strub, F., De~Vries, H., Dumoulin, V., Courville, A.: Film: Visual
  reasoning with a general conditioning layer. In: Proc. of AAAI (2018)

\bibitem{rohrbach2016grounding}
Rohrbach, A., Rohrbach, M., Hu, R., Darrell, T., Schiele, B.: Grounding of
  textual phrases in images by reconstruction. In: Proc. of ECCV (2016)

\bibitem{rupprecht2018guide}
Rupprecht, C., Laina, I., Navab, N., Hager, G.D., Tombari, F.: Guide me:
  Interacting with deep networks. In: Proc. of CVPR (2018)

\bibitem{russakovsky2015imagenet}
Russakovsky, O., Deng, J., Su, H., Krause, J., Satheesh, S., Ma, S., Huang, Z.,
  Karpathy, A., Khosla, A., Bernstein, M., et~al.: Imagenet large scale visual
  recognition challenge. International Journal of Computer Vision
  \textbf{115}(3),  211--252 (2015)

\bibitem{strub2017end}
Strub, F., De~Vries, H., Mary, J., Piot, B., Courville, A., Pietquin, O.:
  End-to-end optimization of goal-driven and visually grounded dialogue systems
  harm de vries. In: Proc. of IJCAI (2017)

\bibitem{sukhbaatar2015end}
Sukhbaatar, S., Weston, J., Fergus, R., et~al.: End-to-end memory networks. In:
  Proc. of NIPS (2015)

\bibitem{devries2017modulating}
de~Vries, H., Strub, F., Mary, J., Larochelle, H., Pietquin, O., Courville,
  A.C.: Modulating early visual processing by language. In: Proc. of NIPS
  (2017)

\bibitem{weston2014memory}
Weston, J., Chopra, S., Bordes, A.: Memory networks. arXiv preprint
  arXiv:1410.3916  (2014)

\bibitem{xiong2016dynamic}
Xiong, C., Merity, S., Socher, R.: Dynamic memory networks for visual and
  textual question answering. In: Proc. of ICML (2016)

\bibitem{xu2016ask}
Xu, H., Saenko, K.: Ask, attend and answer: Exploring question-guided spatial
  attention for visual question answering. In: Proc. of ECCV (2016)

\bibitem{xu2015show}
Xu, K., Ba, J., Kiros, R., Cho, K., Courville, A., Salakhudinov, R., Zemel, R.,
  Bengio, Y.: Show, attend and tell: Neural image caption generation with
  visual attention. In: Proc. of ICML (2015)

\bibitem{yang2018osmn}
Yang, L., Wang, Y., Xiong, X., Yang, J., Katsaggelos, A.K.: Efficient video
  object segmentation via network modulation. In: Proc. of CVPR (2018)

\bibitem{yu2018mattnet}
Yu, L., Lin, Z., Shen, X., Yang, J., Lu, X., Bansal, M., Berg, T.L.: Mattnet:
  Modular attention network for referring expression comprehension. In: Proc.
  of CVPR (2018)

\bibitem{yu2016modeling}
Yu, L., Poirson, P., Yang, S., Berg, A.C., Berg, T.L.: Modeling context in
  referring expressions. In: Proc. of ECCV (2016)

\bibitem{yu2017joint}
Yu, L., Tan, H., Bansal, M., Berg, T.L.: A joint speakerlistener-reinforcer
  model for referring expressions. In: Proc. of CVPR (2016)

\bibitem{zhu2017interactive}
Zhu, Y., Zhang, S., Metaxas, D.: Reasoning about fine-grained attribute phrases
  using reference games. In: Visually-Grounded Interaction and Language
  Workshop (NIPS) (2017)

\bibitem{zhuang2017parallel}
Zhuang, B., Wu, Q., Shen, C., Reid, I.D., van~den Hengel, A.: Parallel
  attention: {A} unified framework for visual object discovery through dialogs
  and queries. Proc. of CVPR  (2018)

\end{thebibliography}

\newpage
\section*{Additional Results}

\subsection*{\Referit ImageClef}

\begin{table}[h]
      \caption{\Referit Guesser Test Error.}
      \label{tab:referit_refclef}
      \centering
       \scriptsize
\begin{tabular}{l|x{2cm}}
        \toprule
        \textbf{Referit} & RefClef\\
        & (berkeley)\\
         & Test \\
        \midrule
        SCRC~\cite{hu2016natural}                & 72.7\%\\
        \midrule  
        Baseline NN+MLB                             & 74.6\%       \\
        \sfilm                                   &84.0\% \\
        \memfilmshort                            &84.3\% \\
        \memfilmshort+(img)                      &$\bm{85.1\%}$ \\
        \bottomrule
        \end{tabular}
\end{table}

\subsection*{Oracle (Without Category Label)}

\begin{table*}[h]
    	\caption{\GW Oracle Test Error without Object Category Label.}
        \label{tab:oracle_results_nocat}
        \centering
        \scriptsize
		\begin{tabular}{l|x{1.1cm}x{1.1cm}x{1.1cm}x{1.1cm}x{1.1cm} | x{ 1.5cm} }
        \toprule
        \textbf{Oracle Model} & Quest. & Dial.  & Spat.  & Image   & Crop   & Test Error \\
        \midrule
        Baseline NN+MLB          & \xmark & \cmark & \cmark    & \cmark  & \cmark & 26.7\%     \\
        \sfilm 			      & \xmark & \cmark & \cmark    & \cmark  & \cmark & 19.5\%     \\
        \memfilmshort 		  & \xmark  & \cmark & \cmark   & \cmark  & \cmark & 18.9\%      \\
        \memfilmshort (+img) & \xmark  & \cmark & \cmark   & \cmark  & \cmark & \bf{18.4\%} \\
        \bottomrule     
        \end{tabular}
\end{table*}

\subsection*{Guesser (Without Category Label)}
\begin{table*}[h]
 \centering
 	   \caption{\GW Guesser Test Error without Object Category Label.}
       \label{tab:guesser_results_nocat}
       \scriptsize
\begin{tabular}{l|x{1.5cm}x{1.5cm}x{1.5cm}}
        \toprule
        \textbf{Guesser Model} & {Crop} & {Image} & {Crop/Img} \\
        \midrule
        PLAN~\cite{zhuang2017parallel}&-& -      & 40.3\% \\
        \memfilmshort          & 35.3\% & 39.8\% & 33.9\% \\
        \memfilmshort (+img)   & 34.3\% & 40.1\% & \bf{33.2}\%\\
        \bottomrule
        \end{tabular}        
\end{table*}
\newpage

\section*{Guesser (Pointing Task)}
\label{sec:pointing}

\begin{table*}[h]
      \centering
       \scriptsize
\caption{Guesser pointing error for different IoU thresholds.}
\begin{tabular}{l|x{1.2cm}x{1.2cm}x{1.45cm}}
        \toprule
        \textbf{Guesser Model} & {IoU $>$ 0.3} & {IoU $> $0.5} & {IoU $>$ 0.7} \\
        \midrule
        Baseline NN       & 81.4\% & 92.0\% & 98.2\% \\
        FiLM              & 74.0\% &85.9\% & 94.7\%\\
        \memfilmshort  & 73.4\% & 84.6\% & 93.7\% \\
        \memfilmshort (+img) & \textbf{71.9\%} &\textbf{84.0\%} & \textbf{93.6\%} \\
        \bottomrule
\end{tabular}
\label{tab:pointing_res}
\end{table*}

For existing tasks on the \GW dataset, the Guesser selects its predicted target object from among a provided list of possible answers. A more natural task would be for the Guesser to directly point out the object, much as a human might. Thus, we introduce a pointing task as a new benchmark for \GW. The specific task is to locate the intended object based on a series of questions and answers; however, instead of selecting the object from a list, the Guesser must output a bounding box around the object of its guess, making the task more challenging.
This task also does not include important side information, namely object category and (x,y)-position~\cite{de2017guesswhat}, making the object retrieval more difficult than the originally introduced Guesser task as well.
The bounding box is defined more specifically as the 4-tuple ($x$, $y$, width, height), where $(x,y)$ is the coordinate of the top left corner of the box within the original image $\Image$, given an input dialogue.

We assess bounding box accuracy using the Intersection Over Union (IoU) metric: the area of the intersection of predicted and ground truth bounding boxes, divided by the area of their union. Prior work~\cite{everingham2010pascal,girshick2014rich}, generally considers an object found if IoU exceeds 0.5. 
\begin{equation}
\textbf{IoU} = \frac{|\text{bboxA}\cap\text{bboxB}| }{| \text{bboxA}\cup\text{bboxB}|} = \frac{ | \text{bboxA}\cap\text{bboxB}| }{ | \text{bboxA}|   + | \text{bboxB}|  -  |\text{bboxA}  \cap \text{bboxB}|}
\end{equation}

We report model error in Table~\ref{tab:pointing_res}. Interestingly, the baseline obtains 92.0\% error while \memfilmshort obtains 84.0\% error. As previously mentioned, re-injecting visual features into the \memfilm's context cell is beneficial. The error rates are relatively high but still in line with those of similar pointing tasks such as SCRC~\cite{hu2016segmentation,hu2016natural} (around 90\%) on \Referit.

\newpage

\begin{figure}[ht]
\section*{Attention Visualizations}
\label{app:mask}
\centering
\includegraphics[width=0.58\linewidth]{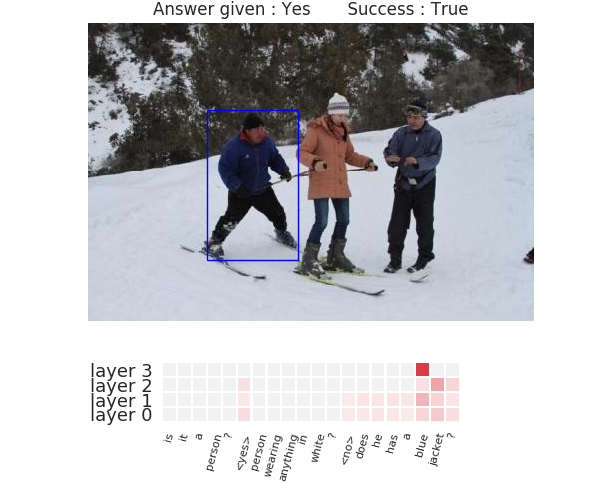}
\includegraphics[width=0.58\linewidth]{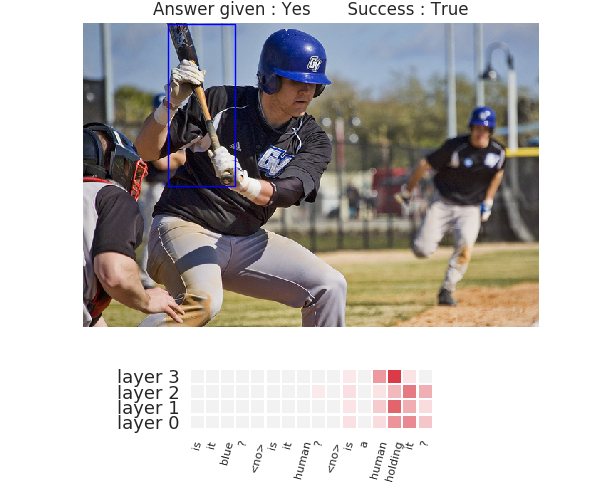}
\includegraphics[width=0.58\linewidth]{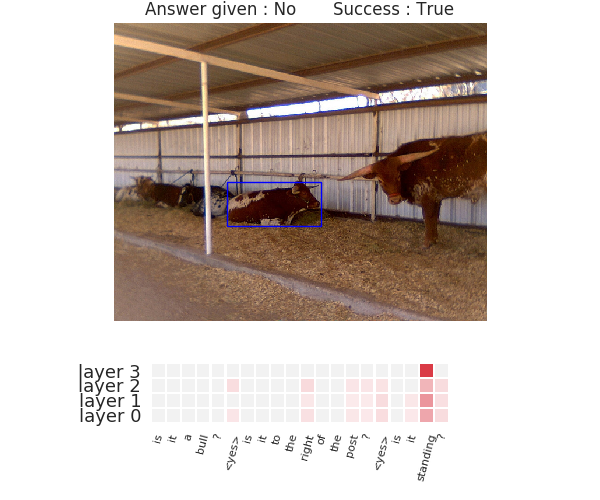}
\caption{The crop pipeline Oracle's attention over the last question when the model succeeds.}
\label{fig:oracle_classic}
\end{figure}

\begin{figure}[ht]
\centering
\includegraphics[width=0.60\linewidth]{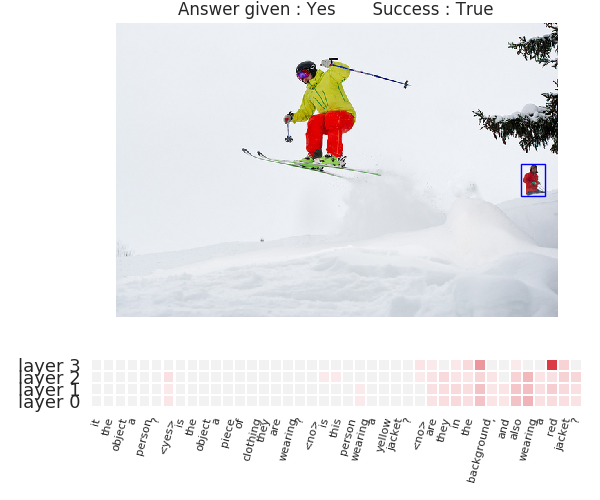}
\includegraphics[width=0.60\linewidth]{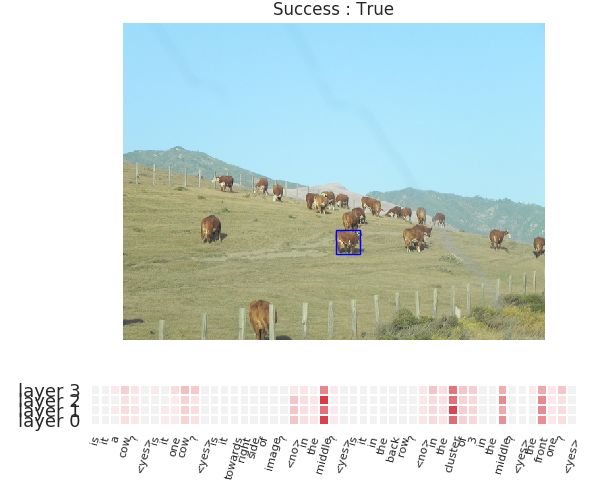}
\includegraphics[width=0.60\linewidth]{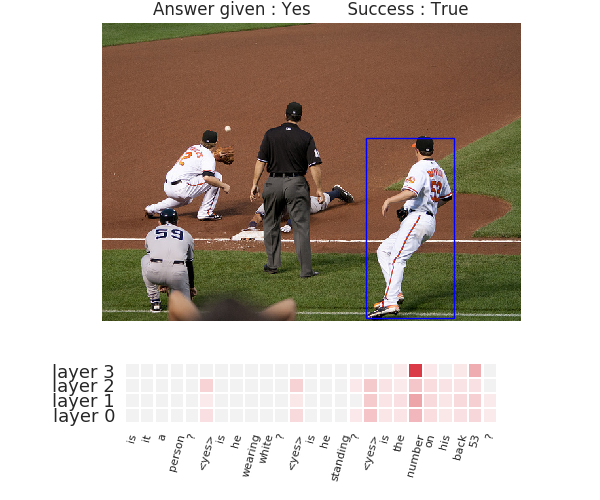}
\caption{The crop pipeline Oracle's attention over the last question, showing more advanced reasoning.}
\label{fig:oracle_complex}
\end{figure}

\begin{figure}[ht]
\centering
\includegraphics[width=0.60\linewidth]{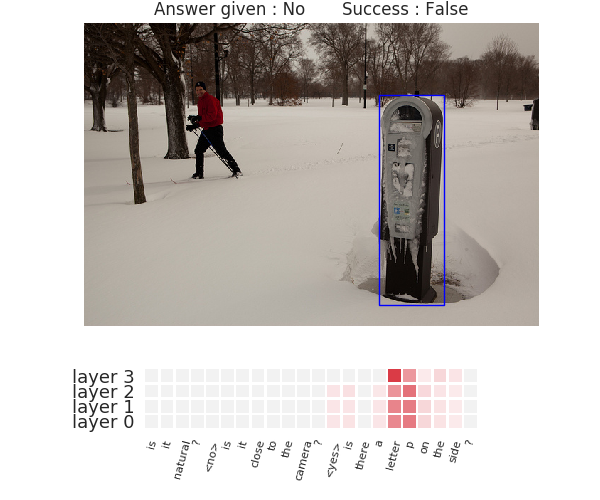}
\includegraphics[width=0.60\linewidth]{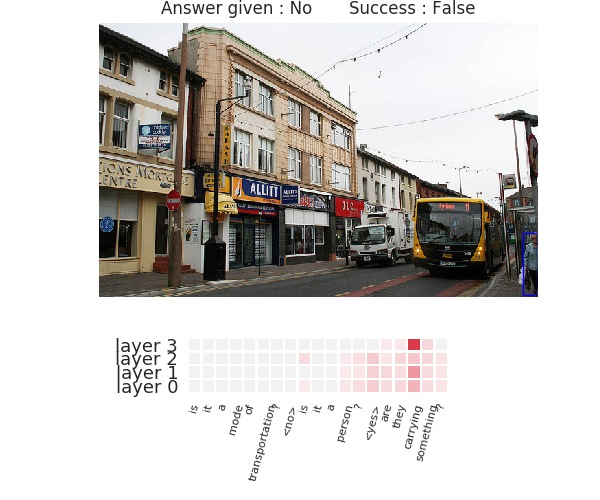}
\includegraphics[width=0.60\linewidth]{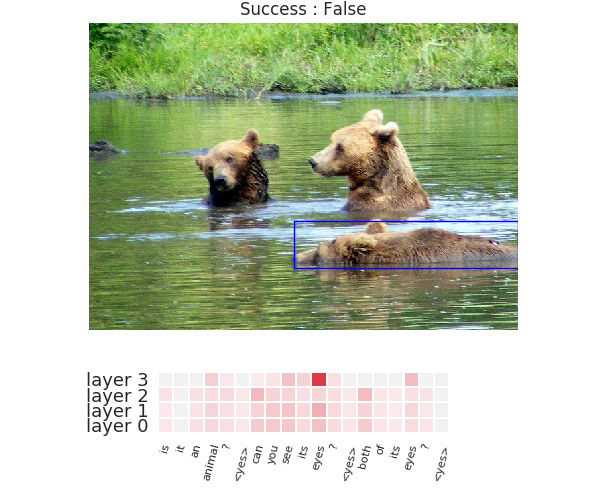}
\caption{The crop pipeline Oracle's attention over the last question when the model fails.}
\label{fig:oracle_negative}
\end{figure}

\begin{figure}[ht]
\centering
\includegraphics[width=0.60\linewidth]{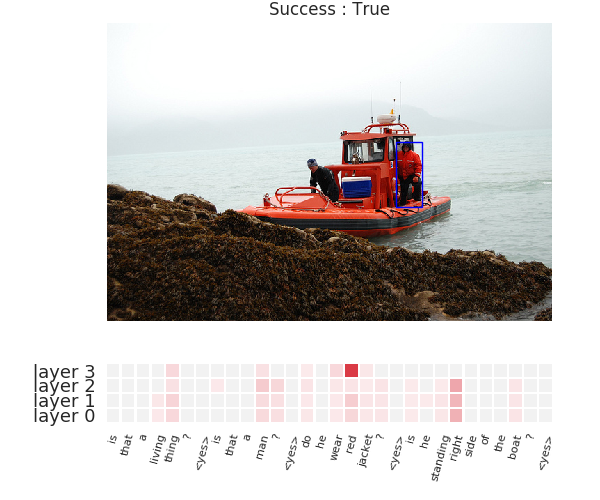}
\includegraphics[width=0.60\linewidth]{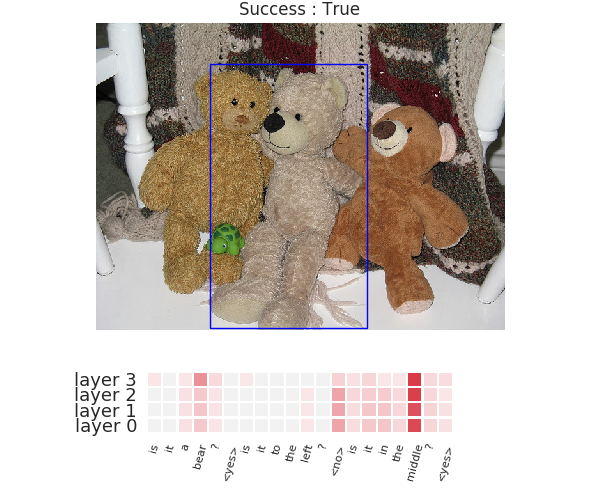}
\includegraphics[width=0.60\linewidth]{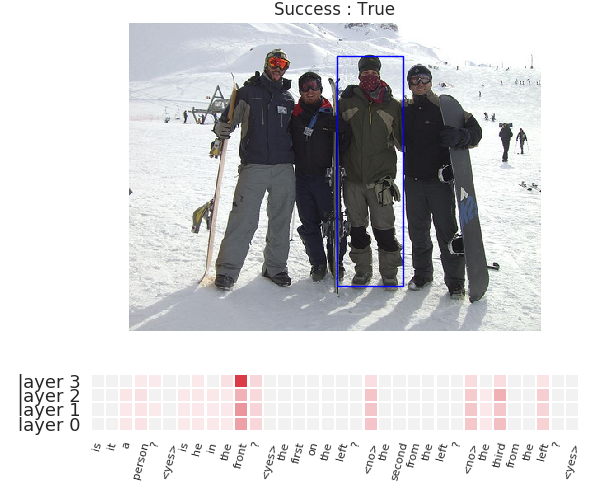}
\caption{The crop pipeline Guesser's attention when the model succeeds.}
\label{fig:guesser_positive}
\end{figure}

\begin{figure}[ht]
\centering
\includegraphics[width=0.60\linewidth]{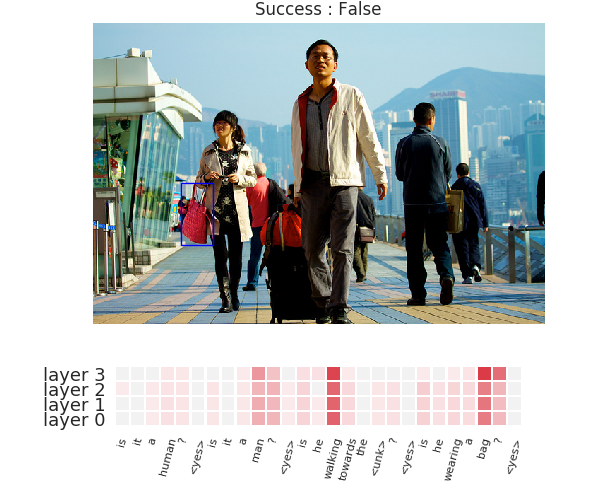}
\includegraphics[width=0.60\linewidth]{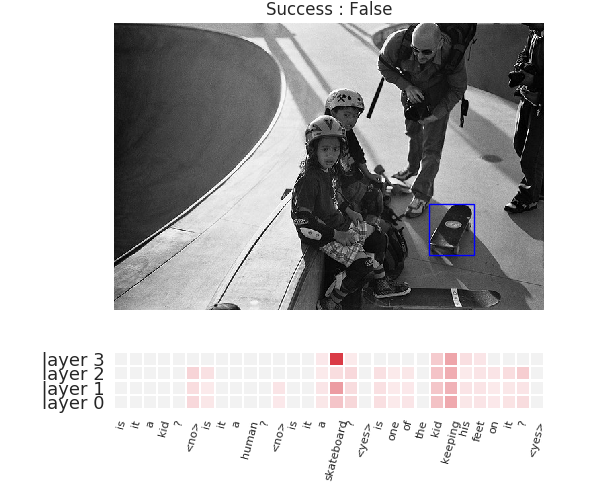}
\includegraphics[width=0.60\linewidth]{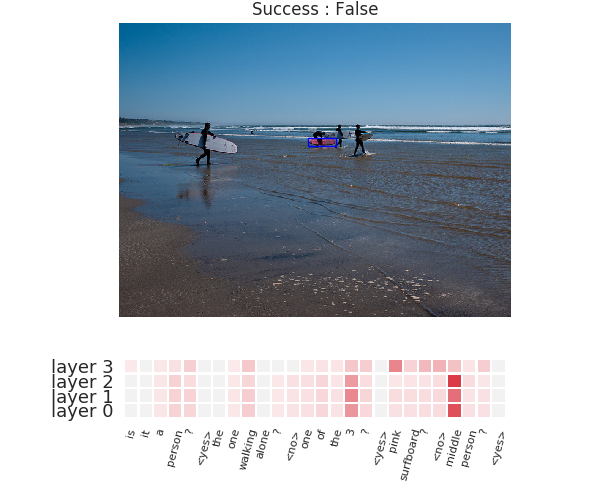}
\caption{The crop pipeline Guesser's attention when the model fails.}
\label{fig:guesser_negative}
\end{figure}

\end{document}